\documentclass[compactaffiliation]{Interspeech}

\interspeechcameraready

\title{Optimizing ASR for Catalan-Spanish Code-Switching: A Comparative Analysis of Methodologies}

\author[affiliation={1}]{Carlos}{Mena}
\author[affiliation={1}]{Pol}{Serra}
\author[affiliation={1}]{Jacobo}{Romero}
\author[affiliation={1}]{Abir}{Messaoudi}
\author[affiliation={1}]{Jose}{Giraldo}
\author[affiliation={1}]{Carme}{Armentano-Oller}
\author[affiliation={1}]{Rodolfo}{Zevallos}
\author[affiliation={2}]{Ivan}{Meza}
\author[affiliation={1,3}]{Javier}{Hernando}

\affiliation{LangTech Lab}{Barcelona Supercomputing Center}{Spain}
\affiliation{IIMAS}{Universidad Nacional Autónoma de México}{Mexico}
\affiliation{}{Universitat Politecnica de Catalunya}{Spain}

\email{carlos.hernandez@bsc.es,ivanvladimir@turing.iimas.unam.mx,javier.hernando@upc.edu}


\usepackage{comment}
\usepackage[utf8]{inputenc} 

\begin{document}

\maketitle
\begin{abstract}



Code-switching (CS), the alternating use of two or more languages, challenges automatic speech recognition (ASR) due to scarce training data and linguistic similarities. The lack of dedicated CS datasets limits ASR performance, as most models rely on monolingual or mixed-language corpora that fail to reflect real-world CS patterns. This issue is critical in multilingual societies where CS occurs in informal and formal settings. A key example is Catalan-Spanish CS, widely used in media and parliamentary speeches. In this work, we improve ASR for Catalan-Spanish CS by exploring three strategies: (1) generating synthetic CS data, (2) concatenating monolingual audio, and (3) leveraging real CS data with language tokens. We extract CS data from Catalan speech corpora and fine-tune OpenAI’s Whisper models, making them available on Hugging Face. Results show that combining a modest amount of synthetic CS data with the dominant language token yields the best transcription performance.

\end{abstract}

\section{Introduction}
\label{sec:into}

Code-switching (CS), also referred to as \textit{code-mixing}, \textit{language alternation}, or various other terms in the literature~\cite{woolard2004codeswitching,hall2015code}, is a phenomenon in which bilingual or multilingual speakers alternate between two or more languages within a conversation, a sentence, or even within a single word~\cite{poplack2013sometimes}. It is a common linguistic practice in communities where multiple languages coexist, often influenced by social, cultural, and cognitive factors~\cite{woolard2004codeswitching}. Code-switching can occur for various reasons, such as emphasizing a point, addressing different audiences, or filling lexical gaps~\cite{hughes2006code}. According to various authors, we can distinguish between three types of alternations ~\cite{ruba2023types,poplack2013sometimes}:

\begin{itemize}
    \item \textbf{Inter-sentential:} Switching occurs between entire sentences, typically at clause boundaries, where each sentence remains grammatically independent in its respective language. (e.g., ``Mañana tengo un examen muy difícil. \emph{I hope I do well}.")  
      
    \item \textbf{Intra-sentential:} Switching happens within a single sentence, mixing words or phrases from different languages while maintaining grammatical coherence. (e.g., ``Ayer fuimos al \emph{mall} y compramos mucha ropa \emph{on sale}.")  
      
    \item \textbf{Tag-switching:} A discourse marker, filler, or tag phrase from another language is inserted into the sentence without affecting its grammatical structure. (e.g., ``Eso es increíble, \emph{you know}?")  
\end{itemize}

In this paper, we explore different approaches to enhancing Automatic Speech Recognition (ASR) in code-switching environments between Catalan and Spanish, which is commonly found in everyday conversations in Catalan-speaking areas. This task is particularly challenging due to the linguistic similarities between both languages. 

Catalan-Spanish code-switching has been widely studied from various perspectives. The literature includes studies on the behavior and acquisition of Catalan-Spanish CS in early childhood~\cite{aguilar2024chapter,aguilar2015code,caballero2022code}, as well as research exploring the influence of cultures from outside Catalonia~\cite{ali2023constructing,hall2015code}, the impact of Catalan culture in other regions of the world~\cite{arnaus2022role,curco2005code}, historical perspectives on CS in different periods of history~\cite{argenter2001code}, and its presence in various domains, such as comedy~\cite{woolard1995changing,woolard1988codeswitching}. More recently, studies on this topic have been conducted in the context of Catalan television~\cite{millan2020social} and social media~\cite{perez2022mixing}.  

However, despite the extensive literature on the subject, there are surprisingly few, if any, speech datasets exclusively dedicated to Catalan-Spanish CS. For this reason, in Section~\ref{sec:methodology}, we include a detailed explanation of how we detect and extract CS utterances from some of the datasets presented in Section~\ref{sec:datasets}. This is a direct contribution of our work to the community, as these subcorpora are publicly shared through the Hugging Face platform.

\section{Datasets}
\label{sec:datasets}

In the name of reproducibility in science, all the data presented in this section is publicly available on the Hugging Face platform. For a quick overview, Table~\ref{tab:datasets} summarizes the datasets presented in this section, the amount of CS speech extracted from them, and their primary function (test or fine-tune).

\begin{table}[ht]
  \caption{Summary of the datasets and their primary function (test or fine-tune).}
  \label{tab:datasets}
  \centering
  \begin{tabular}{lll}
    \toprule
    \textbf{Dataset}      & \textbf{CS Speech} & \textbf{Usage}    \\
    \midrule
    ParlamentParla           & 2 hours    &     Test               \\
    Corts Valencianes        & 2 hours    &     Test               \\
    TV3 Parla                & 1h45m      &     Test               \\
    TV3 Parla                & 4 hours    &     fine-tune           \\
    Common Voice c17.0       & 105 hours  &     fine-tune           \\
    Synthetic CS Data        & 17 hours   &     fine-tune           \\
    \bottomrule
  \end{tabular}
\end{table}

\subsection{ParlamentParla}
\label{subsec:parla}

ParlamentParla~\cite{kulebi_baybars_2021_5541827} is a speech corpus for Catalan, prepared by Col·lectivaT~\cite{collectivat2024}, consisting of audio segments extracted from Catalan Parliament plenary sessions (2007–2018). The transcriptions were aligned with the recordings, and the data is released under the Parliament's terms of use\footnote{\url{https://www.parlament.cat/pcat/serveis-parlament/avis-legal/}}. 

Currently, no in-depth linguistic study of the corpus has been conducted. However, we assume that the predominant accent is Central Catalan, as it corresponds to the geographical area where most of Catalonia's population resides~\cite{feldhausen2010sentential}.

The corpus was supported by the Catalan Department of Culture and, in v2.0, by the Barcelona Supercomputing Center through the AINA project~\cite{parlamentparla2021huggingface}. Version 2.0~\cite{parlamentparla2021github} includes 211 hours of clean speech and 400 hours of other quality segments, with speaker and gender tags.

\subsection{Corts Valencianes}

The Corts Valencianes Speech Corpus~\cite{bscib32024} is a comprehensive dataset consisting of speech recordings from the sessions of the Corts Valencianes\footnote{\url{https://www.cortsvalencianes.es/}}. It features high-quality and other-quality segments, categorized into short segments (under 30 seconds) and long segments (over 30 seconds). The dataset contains a total of 2,621,096 words and covers a total of 270 of speech, with 239 hours dedicated to short segments and 31 hours to long segments.

Although no in-depth linguistic study has been conducted in this case, it can be inferred that the predominant accent is Valencian, as it is the most common in this region.

\subsection{TV3 Parla}

TV3 Parla~\cite{kulebi2021tv3parla} is a 272-hour Catalan corpus extracted from the Catalan public television channel TV3~\cite{quinta1983tv3}. The details of segmentation and data processing are explained in~\cite{kulebi2018building}. Its content is publicly available but owned by Corporació Catalana de Mitjans Audiovisuals, SA (CCMA).

As in Section~\ref{subsec:parla}, we assume for demographic reasons that the predominant accent is the Central variety, albeit to a lesser extent, as one of the objectives of Catalan public television is to promote the Catalan language in all its varieties.

\subsection{Mozilla Common Voice v17.0}

Mozilla CommonVoice~\cite{commonvoice2020} is a collaborative project aimed at creating an open-source dataset of voice recordings to improve speech recognition technologies, especially for underrepresented languages. The project collects speech data from volunteers worldwide, ensuring diversity in accents, demographics, and speaking styles.

In our experiments, we use version 17 of the Common Voice corpus~\cite{commonvoice17} for both Spanish and Catalan, leveraging its high-quality transcriptions and diverse speaker contributions to enhance our ASR models.

\subsection{Synthetic CS Data}

The synthetic audio corpus comprises 17 hours of CS data, with sentences extracted from the TV3 Parla corpus through a process detailed in Section~\ref{sec:syntdatagen}. In essence, this data is generated by taking a Catalan text corpus (TV3 Parla) and introducing code-switching by translating noun chunks into Spanish, ensuring a natural integration within the original linguistic context. Once the text corpus was prepared, speech was synthesized using TTS technology.

\section{Methodology}
\label{sec:methodology}

This work aims to test different methods for improving the performance of ASR systems in Catalan and Spanish CS contexts. The first method involves generating synthetic CS sentences and then synthesizing them using a Text-to-Speech (TTS) system, while the second method consists of concatenating audio recordings in both languages that contain real human voices. The third method implies the detection of language tokens in transcriptions.

\begin{table*}[ht]
\centering
\caption{WER (\%) of different fine-tuning and decoding strategies. The experiments were done on the Code-switching parts of the datasets.}
\label{tab:results1}
\begin{tabular}{lcccc}
\hline
\textbf{Experiment}     & \textbf{Decoding Tokens}      & \textbf{TV3}   & \textbf{ParlamentParla} & \textbf{Corts} \\ 
\hline
Base model     &                        & 31.96 & 20.96         & 35.43 \\
               & $<$ca$><$es$>$         & 25.12 & 22.39         & 29.81 \\
               & $<$es$><$ca$>$         & 25.49 & 20.13         & 34.16 \\
               & $<$ca$>$               & 26.67 & 22.17         & 30.34 \\
               & $<$es$>$               & 63.16 & 57.87         & 52.16 \\
\hline
Synthetic data & $<$ca$><$es$>$         & 23.22 & 16.38         & 22.82 \\
               & $<$es$><$ca$>$         & 23.81 & 15.47         & 22.71 \\
               & $<$ca$>$               & 23.46 & \textbf{14.48} & \textbf{22.42} \\
               & $<$es$>$               & 24.23 & 15.72         & 23.76 \\
\hline
TV3            & $<$ca$><$es$>$         & 21.96 & 15.76         & 23.98 \\
               & $<$es$><$ca$>$         & 21.46 & 16.65         & 24.86 \\
               & $<$ca$>$               & \textbf{21.20} & 16.01& 24.22 \\
               & $<$es$>$               & 51.29 & 50.58         & 47.75 \\
\hline
Tuples         & $<$ca$><$es$>$         & 28.99 & 17.83         & 25.24 \\ 
               & $<$es$><$ca$>$         & 29.92 & 21.37         & 29.72 \\ 
               & $<$ca$>$               & 31.42 & 18.31         & 26.06 \\ 
               & $<$es$>$               & 52.78 & 50.22         & 47.32 \\ 
\hline
\end{tabular}
\end{table*}

\subsection{Creation of the Evaluation Dataset}
\label{sec:creation-eval-dataset}

To obtain gold standard data to evaluate the effectiveness of our ASR models, we followed two main strategies: manual and BERT~\cite{kenton2019bert} detection.

\subsubsection{Manual Detection of CS}

For the manual detection, we extracted CS sentences from the \textit{ParlamentParla} and \textit{Corts Valencianes} corpora. These datasets primarily contain Catalan speech in a parliamentary discourse context, delivered by Catalan and Valencian politicians. The method used to detect CS sentences was keyword-based detection. For example, in Catalan, the conjunction ``and" is ``i", while in Spanish, it is ``y". Therefore, sentences containing ``y" are likely to include CS. Once sentences were detected using this and other keywords, they were manually verified, and only those containing at least three consecutive words in Spanish were accepted to be part of the evaluation set.

\subsubsection{BERT Detection of CS}

In the case of the BERT detection, Google’s multilingual BERT~\cite{devlin-etal-2019-bert} was fine-tuned for token classification using a synthetic corpus of code-switched dialogues in Catalan and Spanish. During fine-tuning, each word was labelled with its corresponding language token. Once trained, the model was applied to the transcriptions of the \textit{TV3 Parla} dataset, where it performed token-level language classification. This process resulted in a ``language count" for each audio file, indicating the distribution of Catalan and Spanish within the transcription. Given that the audios were short, an audio was considered code-switched if both Catalan and Spanish were present with at least three words each. With this method, we identified a substantial portion of code-switched data, totalling approximately 5 hours and 45 minutes. This data was then split for model training, with 70\% used for fine-tuning and 30\% for validation and testing.


\subsection{Synthetic CS Generation}
\label{sec:syntdatagen}

The creation of synthetic corpora involved two main processes: text generation and audio generation through text-to-speech (TTS). To generate the synthetic text corpus with code-switching between Catalan and Spanish, sentences were first extracted from the TV3 Parla ~\cite{kulebi2021tv3parla}, filtering them based on their length. Then, one or two \textit{noun chunks} per sentence were identified using \texttt{spaCy}~\cite{honnibal2017spacy} and translated into Spanish with \texttt{MarianMT}~\cite{huggingface_marian}, replacing the original segments with \texttt{<cat>} and \texttt{<esp>} tags.

To synthesize the generated CS text, we trained a multilingual TTS based on a Catalan Matcha-TTS~\cite{mehta2024,peirolilja24interspeech} version. The datasets used for the TTS training were Librivox Spanish~\cite{carlosmena2020librivoxspanish}, Festcat~\cite{bonafonte-etal-2008-corpus} and LaFresCat~\cite{peirolilja24_iberspeech}. We added an embedding for language and a custom front-end to handle CS text. As the synthetic text was generated with language tags, it allowed us to route the sentences to the proper phonemizer together with the language index of the embedding. The whole text corpus was synthesized with 10 voices rated the best in an internal listening assessment. The final step involved a data augmentation pipeline with the audiomentations\footnote{\url{https://github.com/iver56/audiomentations} 
    } library since the goal was to simulate a real user environment instead of the clean that outputs the TTS. This pipeline used the following transformations with randomization of the parameters:

\begin{itemize}
    \item Additive environmental noise and speech babble noise. $(min\_absolute\_rms\_in\_db=-45.0, max\_absolute\_rms\_in\_db=-40.0)$
    \item Clipping $(a\_min=-0.07, a\_max=0.07, p=1.0)$
    \item Tanh distortion $(min\_distortion=0.3, max\_distortion=0.3, p=0.5)$
    \item Gradual gain transition $(min\_gain\_db=-3, max\_gain_db=3, min\_duration=0.5, max\_duration=1, p=1.0)$
    \item Bitcrushing $(min\_bit\_depth=8, max\_bit\_depth=8, p=1.0)$
\end{itemize}


\subsection{Audio Concatenation}
\label{sec:audioconcat}
The idea of concatenating audio from different languages as a technique to handle CS in ASR is not new at all. The NVIDIA-NeMo GitHub repository provides scripts to create synthetic code-switched data using this approach\footnote{\url{https://github.com/NVIDIA/NeMo/tree/bc4bce71d01234f568c1327f0848001d86143b3d/scripts/speech_recognition/code_switching}}. However, in this work, we implemented our version of this script to ensure that the concatenation was performed exactly as intended.

The Spanish and Catalan audio files used for concatenation were extracted from version 17 of Mozilla Common Voice. We selected 50 hours per language, ensuring that the speakers' gender was well specified (male or female) to create a gender-balanced corpus.

Audio files were concatenated in pairs (tuples), ensuring that each audio segment in one language was always paired with another segment in the other language. If 30,000 audio clips (50 hours) from one language are concatenated with 30,000 clips (50 hours) from the other, the result is 30,000 tuples (100 hours). Of these, 15,000 will be of type ca-es (starting with Catalan and then Spanish), and 15,000 will be of type es-ca (starting with Spanish and then Catalan).

\subsection{Language Tag Detection}

Language tag detection is a technique that has demonstrated acceptable results when used with other languages~\cite{yang2024adapting}. We carried out a series of experiments to identify the most effective strategy for handling CS between Catalan and Spanish while minimizing transcription errors. Our approach builds on the original configuration of Whisper, which employs the following prompt: \texttt{<|startofprev|>previous text<|startoftranscript|><|language|>} \texttt{<|task|><|notimestamps|>}, as described in \cite{peng2023prompting}. We modified this setup to assess its impact on CS transcription performance.

Specifically, we investigated the effect of concatenating language tags on transcription accuracy for CS utterances. This involved explicitly enforcing two language tags (e.g., \texttt{<|ca|><|es|>}) within the prompt: \texttt{<|startoftranscript|><|ca|><|es|><|task|>} \texttt{<|notimestamps|>}. Previous findings from \cite{peng2023prompting} reported a 19\% 

Additionally, we examined the influence of varying the position of language tags in the decoding sequence by placing Catalan first, Spanish first, or using a single language tag (either Spanish or Catalan).

\section{Experiments}
\label{sec:experiments}


We performed multiple fine-tuning and decoding experiments to evaluate different strategies for handling Catalan-Spanish CS. Our objective was to identify the most effective approach for minimizing transcription errors. To achieve this, we fine-tuned Whisper Large v3~\cite{whisperlargev3} on three different datasets, as shown in Table~\ref{tab:results1}.

Table~\ref{tab:results1} presents the results of our fine-tuning experiments.  The ``Base model" refers to the original Whisper model without additional fine-tuning. The ``Synthetic data" approach involves training on artificially generated Catalan-Spanish CS speech, as described in Section~\ref{sec:syntdatagen}.  The ``TV3" dataset consists of real-world broadcast speech containing natural CS instances, making it a strong candidate for domain adaptation after fine-tuning.  Finally, the ``Tuples" dataset (Section~\ref{sec:audioconcat}) is built from read speech of multiple native speakers, and its structure encourages model generalization across different CS patterns.

In the decoding phase, we investigated multiple tokenization strategies to determine their impact on transcription accuracy. Specifically, we experimented with dual-language tokens (\texttt{<ca><es>} and \texttt{<es><ca>}), which explicitly specify both languages in the sequence. We also tested single-language tokens (\texttt{<ca>} or \texttt{<es>}), where only one language is enforced during decoding. Finally, we evaluated the model’s behavior when no explicit language token was provided.

\section{Analysis of the Results}
\label{sec:results}

Table \ref{tab:results1} presents the WER scores across three test sets (TV3, ParlamentParla, and Corts). Several key observations emerge from the results. First, fine-tuning significantly improves recognition accuracy compared to the base model across all datasets.

The Synthetic Data approach yields competitive results, particularly for ParlamentParla (14.48\% WER), highlighting the potential of synthetic augmentation for domain adaptation.

The model fine-tuned on TV3 achieves the best performance only on the TV3 test set (21.20\% WER), implying that it does not adapt well to other real-world CS scenarios. When the 4 rows corresponding to the TV3 results are removed, the best result is obtained by the Synthetic Data approach (23.22\% WER). 

Conversely, the Tuples-based model does not generalize as effectively, showing higher WER than both the TV3 and synthetic models. This may be because randomly concatenating audio segments introduces inconsistencies within the transcript, negatively impacting the ASR model’s performance as it relies on contextual information for transcription.

In terms of decoding strategies, the ordering of the language tags (\texttt{<ca><es>} vs. \texttt{<es><ca>}) has minimal impact, as both configurations yield comparable results. However, the best results are consistently obtained with the language token \texttt{<ca>}.

In contrast, the language token \texttt{<es>} consistently leads to substantial performance degradation, with WER increasing dramatically to 63.16\% in the Base model and 51.29\% in the TV3 model on the TV3 test set. This suggests that the model struggles when forced to assume Spanish transcription in a Catalan-prevalent CS scenario.

It is also worth noting that the Synthetic Data approach provides the best results using only 17 hours of data, compared to the 100 hours of read speech in the Tuples approach or the 4 hours of spontaneous speech from TV3. This is a remarkable finding, which warrants further exploration for other languages in future experiments.

However, the 4 hours of audio in the TV3 dataset may not be enough to achieve optimal results, and adding more code-switched data to the training might effectively make this fine-tuned model more generalizable to other datasets. Moreover, integrating the code-switched data in \textit{ParlamentParla}, \textit{Corts Valencianes} and \textit{TV3 Parla} for fine-tuning in future research could potentially lead to even better performance.

\section{Conclusions and Further Work}
\label{sec:conclusions}

In this work, we have presented a broad overview of the current situation of code-switching in the Catalonia region through an extensive literature review. We have described the datasets used for our experiments, which are publicly available to the community via the Hugging Face platform. We detailed the three methodologies explored in this work: generating synthetic code-switched data, concatenating audio recordings in Catalan and Spanish, and leveraging real code-switched data by detecting language tokens in transcriptions. With this in mind, we present a series of results that suggest the best strategy for handling Catalan-Spanish code-switching is to generate a modest amount of synthetic data using the methodology described in Section~\ref{sec:syntdatagen}, and to use the language token for the more prevalent language, in this case, Catalan (\texttt{<ca>}). 

As future work, we aim to apply these findings to creating ASR models and datasets that involve more languages. However, this work represents a modest step in the right direction, and we hope it contributes to the proper assimilation of code-switching by ASR systems of all kinds.

\section{Acknowledgements}


This work is funded by the Ministerio para la Transformación Digital y de la Función Pública and Plan de Recuperación, Transformación y Resiliencia - Funded by EU – NextGenerationEU within the framework of the project ILENIA with reference 2022/TL22/00215337 and by the Government of Catalonia through the Aina project.


\bibliographystyle{IEEEtran}
\bibliography{mybib}

\end{document}